\documentclass[11pt, a4paper, logo, copyright, nonumbering]{deepseek}
\usepackage[authoryear, compress, round]{natbib}
\usepackage{dblfloatfix}
\usepackage{ulem}
\usepackage{caption}
\usepackage{dramatist}
\usepackage{xspace}
\usepackage{pifont}
\usepackage{multirow}
\usepackage{tcolorbox}
\usepackage{xltabular}
\usepackage{longtable}
\usepackage{hyperref}
\usepackage{amsfonts}
\usepackage{amsmath}
\usepackage{amssymb}
\usepackage{lineno}

\usepackage[bottom]{footmisc}

\usepackage{CJKutf8}
\usepackage{subfigure}
\usepackage{setspace}

\usepackage{wrapfig}

\usepackage[dvipsnames]{xcolor}

\definecolor{keywordcolor}{rgb}{0.7, 0.1, 0.1}   
\definecolor{tacticcolor}{rgb}{0.0, 0.1, 0.6}    
\definecolor{commentcolor}{rgb}{0.4, 0.4, 0.4}   
\definecolor{symbolcolor}{rgb}{0.0, 0.1, 0.6}    
\definecolor{sortcolor}{rgb}{0.1, 0.5, 0.1}      
\definecolor{attributecolor}{rgb}{0.7, 0.1, 0.1} 
\usepackage{listings}

\definecolor{mygray}{gray}{0.9}

\makeatletter
\def\@BTrule[#1]{%
  \ifx\longtable\undefined
    \let\@BTswitch\@BTnormal
  \else\ifx\hline\LT@hline
    \nobreak
    \let\@BTswitch\@BLTrule
  \else
     \let\@BTswitch\@BTnormal
  \fi\fi
  \global\@thisrulewidth=#1\relax
  \ifnum\@thisruleclass=\tw@\vskip\@aboverulesep\else
  \ifnum\@lastruleclass=\z@\vskip\@aboverulesep\else
  \ifnum\@lastruleclass=\@ne\vskip\doublerulesep\fi\fi\fi
  \@BTswitch}
\makeatother

\addto\extrasenglish{
}

 {\begin{list}{}%
         {\setlength{\leftmargin}{#1}}%
         \item[]%
 }
 {\end{list}}

\reportnumber{001} 

\renewcommand{\today}{}

\title{\centering \Large DeepSeekMath-V2: Towards Self-Verifiable Mathematical Reasoning}

\author[*]{
\footnotesize\vspace{-0.1in}
Zhihong Shao*, Yuxiang Luo*, Chengda Lu*$^\dag$, Z.Z. Ren*

Jiewen Hu, Tian Ye, Zhibin Gou, Shirong Ma, Xiaokang Zhang

\small
DeepSeek-AI

\small
zhihongshao@deepseek.com \\
\small
\url{https://github.com/deepseek-ai/DeepSeek-Math-V2}
\vspace{-0.2in}
}
\correspondingauthor{Core contributors ~ $^\dag$Work done during internship at DeepSeek-AI.}

\renewcommand{\phi}{\varphi}

\renewcommand{\epsilon}{\varepsilon}
\renewcommand{\imath}{\mathrm{i}}

\newlength{\restsubwidth}
\newlength{\restsubheight}
\newlength{\restsubmoreheight}
\newcommand{\rest}[2]{%
        \settowidth{\restsubwidth}{\ensuremath{#2}}
        \settoheight{\restsubheight}{\ensuremath{{}_{#2}}}
        \ensuremath{{#1\hskip 0.5pt}_{\vrule\kern2pt\parbox[b][%
        4pt][b]{\the\restsubwidth}{%
                        \ensuremath{{}_{#2}}}}}
        }

\usepackage{thmtools}

\begin{abstract}
Large language models have made significant progress in mathematical reasoning, which serves as an important testbed for AI and could impact scientific research if further advanced.
By scaling reasoning with reinforcement learning that rewards correct final answers, LLMs have improved from poor performance to saturating quantitative reasoning competitions like AIME and HMMT in one year.
However, this approach faces fundamental limitations.
Pursuing higher final answer accuracy doesn't address a key issue: correct answers don't guarantee correct reasoning.
Moreover, many mathematical tasks like theorem proving require rigorous step-by-step derivation rather than numerical answers, making final answer rewards inapplicable.
To push the limits of deep reasoning, we believe it is necessary to verify the comprehensiveness and rigor of mathematical reasoning.
Self-verification is particularly important for scaling test-time compute, especially for open problems without known solutions.
Towards self-verifiable mathematical reasoning, we investigate how to train an accurate and faithful LLM-based verifier for theorem proving.
We then train a proof generator using the verifier as the reward model, and incentivize the generator to identify and resolve as many issues as possible in their own proofs before finalizing them.
To maintain the generation-verification gap as the generator becomes stronger, we propose to scale verification compute to automatically label new hard-to-verify proofs, creating training data to further improve the verifier.
Our resulting model, DeepSeekMath-V2, demonstrates strong theorem-proving capabilities, achieving gold-level scores on IMO 2025 and CMO 2024 and a near-perfect 118/120 on Putnam 2024 with scaled test-time compute.
While much work remains, these results suggest that self-verifiable mathematical reasoning is a feasible research direction that may help develop more capable mathematical AI systems.
\end{abstract}

\begin{document}
\begin{CJK*}{UTF8}{gbsn}

\maketitle


\vspace{-0.05in}



\section{Introduction}\label{sec:introduction}

The conventional approach to reinforcement learning (RL) for mathematical reasoning involves rewarding large language models (LLMs) based on whether their predicted final answers to quantitative reasoning problems match ground-truth answers \citep{deepseek-r1}.
This methodology suffices to allow frontier LLMs to saturate mathematical competitions that primarily evaluate final answers, such as AIME and HMMT.
However, this reward mechanism has two fundamental limitations.
First, it serves as an unreliable proxy for reasoning correctness -- a model can arrive at the correct answer through flawed logic or fortunate errors.
Second, it is inapplicable to theorem proving tasks, where problems may not require producing numerical final answers and rigorous derivation is the primary objective.

Consequently, LLMs trained on quantitative reasoning problems with such final answer reward still frequently produce mathematically invalid or logically inconsistent natural-language proofs.
Moreover, this training approach does not naturally develop the models' ability to verify proof validity -- they exhibit high false-positive rates, often claiming incorrect proofs are valid even when they contain obvious logical flaws.

The lack of a generation-verification gap in natural-language theorem proving hinders further improvement.
To address this, we propose developing proof verification capabilities in LLMs.
Our approach is motivated by several key observations:
\begin{itemize}
    \item Humans can identify issues in proofs even without reference solutions -- a crucial ability when tackling open problems.
    \item A proof is more likely to be valid when no issues can be identified despite scaled verification efforts.
    \item The efforts required to identify valid issues can serve as a proxy for proof quality, which can be exploited to optimize proof generation.
\end{itemize}
We believe that LLMs can be trained to identify proof issues without reference solutions.
Such a verifier would enable an iterative improvement cycle: (1) using verification feedback to optimize proof generation, (2) scaling verification compute to auto-label hard-to-verify new proofs, thereby creating the training data to improve the verifier itself, and (3) using this enhanced verifier to further optimize proof generation.
Moreover, a reliable proof verifier enables us to teach proof generators to evaluate proofs as the verifier does.
This allows a proof generator to iteratively refine its proofs until it can no longer identify or resolve any issues.
In essence, we make the model explicitly aware of its reward function and enable it to maximize this reward through deliberate reasoning rather than blind trial-and-error.

Built on DeepSeek-V3.2-Exp-Base \citep{deepseekv32}, we developed \textbf{DeepSeekMath-V2}, a large language model optimized for natural-language theorem proving that demonstrates self-verifiable mathematical reasoning.
Our model can assess and iteratively improve its own proofs, achieving gold-level performance in premier high-school mathematics competitions including IMO 2025 and CMO 2024.
On the Putnam 2024 undergraduate competition, it scored 118/120, exceeding the highest score of 90 \footnote{\url{https://kskedlaya.org/putnam-archive/putnam2024stats.html}} obtained by human participants.

\section{Method}\label{sec:method}

\subsection{Proof Verification}

\subsubsection{Training a Verifier to Identify Issues and Score Proofs}
\label{sec:verifier}

We developed high-level rubrics $\mathcal{I}_v$ for proof evaluation (see Appendix~\ref{app:proof_verification}) with the goal of training a verifier to evaluate proofs according to these rubrics, mirroring mathematical experts' assessment process.
Specifically, given a problem $X$ and a proof $Y$, the verifier $\pi_\phi(\cdot\vert{}X, Y, \mathcal{I}_v)$ is designed to produce a proof analysis that first summarizes identified issues (if any) and then assigns a score based on three levels:
1 for complete and rigorous proofs with all logical steps clearly justified;
0.5 for proofs with sound overall logic but minor errors or omitted details;
and 0 for fundamentally flawed proofs containing fatal logical errors or critical gaps.

\paragraph{Curating Cold Start RL Data}
We constructed our initial training data through the following process:

\begin{enumerate}
    \item We crawled problems from Art of Problem Solving (AoPS) contests \footnote{\url{https://artofproblemsolving.com/community/c13_contest_collections}}, prioritizing math olympiads, team selection tests, and post-2010 problems explicitly requiring proofs, totaling 17,503 problems.
    This problem set is denoted as $\mathcal{D}_p$.
    
    \item We generated candidate proofs using a variant of DeepSeek-V3.2-Exp-Thinking. As this model was not optimized for theorem proving and tended to produce concise but error-prone outputs, we prompted it to iteratively refine its proofs over multiple rounds to improve comprehensiveness and rigor.
    
    \item We randomly sampled proofs across diverse problem types (e.g., algebra and number theory) and had mathematical experts score each proof according to the evaluation rubrics described above.
\end{enumerate}
This process yielded an initial RL dataset $\mathcal{D}_v = \{(X_i, Y_i, s_i)\}$, where each item consists of a problem $X_i$, a proof $Y_i$, and an overall proof score $s_i \in \{0, 0.5, 1\}$.

\paragraph{RL Objective.}  
Building on a version of DeepSeek-V3.2-Exp-SFT which was supervised fine-tuned on reasoning data related to mathematics and code, we trained the model with reinforcement learning to produce proof analyses using two reward components:

\begin{itemize}
    \item \textbf{Format reward} $R_{\text{format}}$: An indicator function that enforces the model to generate both a summary of identified issues and a proof score, by checking whether the final response contains the key phrase ``Here is my evaluation of the solution:'' as well as a score within \texttt{\textbackslash boxed\{\}} following ``Based on my evaluation, the final overall score should be:''.
    
    \item \textbf{Score reward} $R_{\text{score}}$: Rewards based on proximity between predicted score $s_i'$ and annotated score $s_i$:
    \begin{equation}
        R_{\text{score}}(s_i', s_i) = 1 - |s_i' - s_i|
    \end{equation}
\end{itemize}
The RL objective for training the verifier is:
\begin{equation}
    \max_{\pi_\phi} \mathbb{E}_{(X_i, Y_i, s_i) \sim \mathcal{D}_v, (V_i', s_i') \sim \pi_\phi(\cdot|X_i, Y_i)} \left[ R_{\text{format}}(V_i') \cdot R_{\text{score}}(s_i', s_i) \right]
\end{equation}
where $V_i'$ denotes the verifier's final response and $s_i'$ is the proof score extracted from it.

\subsubsection{Introducing Meta-Verification to Review Proof Analyses}

The approach described in Section~\ref{sec:verifier} trains proof verification through RL to align predicted proof scores with expert annotations, but provides no direct supervision on the identified issues themselves.
This creates a critical vulnerability: when evaluating flawed proofs (where $s_i < 1$) during training, the verifier can receive full reward by predicting the correct scores while hallucinating non-existent issues, undermining its trustworthiness.

To address this problem, we introduce \textbf{meta-verification}: a secondary evaluation process that assesses whether issues identified by the verifier indeed exist and whether these issues logically justify the predicted proof score according to the evaluation rubrics $\mathcal{I}_v$.
The complete meta-verification rubrics $\mathcal{I}_{mv}$ are detailed in Appendix~\ref{app:meta_verification}.

We trained a dedicated meta-verifier using RL to perform this evaluation.
By incorporating the meta-verifier's feedback into verifier training, we can improve the faithfulness of the verifier's issue identification.

\paragraph{Meta-Verifier Training Process}

\begin{enumerate}
    \item We obtained an initial verifier $\pi_\phi$ following Section~\ref{sec:verifier}.
    
    \item Mathematical experts scored the quality of verifier responses according to $\mathcal{I}_{mv}$, creating dataset $\mathcal{D}_{mv} = \{(X_i, Y_i, V_i, ms_i)\}$, where $V_i$ is the analysis of proof $Y_i$ and $ms_i \in \{0, 0.5, 1\}$ is the expert-annotated quality score.
    
    \item We trained a meta-verifier $\pi_\eta(\cdot|X, Y, V, \mathcal{I}_{mv})$ to analyze the verifier's proof analysis $V$.
    The meta-verifier produces a summary of issues found in the analysis itself, followed by a quality score measuring how accurate and justified the verifier's analysis is.
    The RL objective follows the same structure as the verifier training, with format and score rewards.

\end{enumerate}

Using the trained meta-verifier $\pi_\eta$, we enhanced the verifier training by integrating meta-verification feedback into the reward function:
\begin{equation}
    R_V = R_{\text{format}} \cdot R_{\text{score}} \cdot R_{\text{meta}}
\end{equation}
where $R_{\text{meta}}$ is the quality score from the meta-verifier.

We trained the enhanced verifier on both the verification dataset $\mathcal{D}_v$ and the meta-verification dataset $\mathcal{D}_{mv}$, using the same reward mechanism on $\mathcal{D}_{mv}$ as used for training the meta-verifier.
The resulting model can perform both proof verification and meta-verification tasks.

On a validation split of $\mathcal{D}_v$, the average quality score of the verifier's proof analyses -- as evaluated by the meta-verifier -- improved from 0.85 to 0.96, while maintaining the same accuracy in proof score prediction.

\subsection{Proof Generation}

\subsubsection{Training a Generator for Theorem Proving}

With verifier $\pi_\phi$ serving as a generative reward model, we train a proof generator $\pi_\theta(\cdot\vert{}X)$ with the RL objective:

\begin{equation}
    \max_{\pi_\theta} \mathbb{E}_{X_i \sim \mathcal{D}_p, Y_i \sim \pi_\theta(\cdot|X_i)} [R_Y]
\end{equation}
where $R_Y$ is the proof score produced by $\pi_\phi(\cdot\vert{}X_i, Y_i, \mathcal{I}_v)$.

\subsubsection{Enhancing Reasoning via Self-Verification}

When a proof generator fails to produce a completely correct proof in one shot -- common for challenging problems from competitions like IMO and CMO -- iterative verification and refinement can improve results.
This involves analyzing the proof with an external verifier and prompting the generator to address identified issues.

However, we observed a critical limitation: when prompted to both generate and analyze its own proof in one shot, the generator tends to claim correctness even when the external verifier easily identify flaws.
In other words, while the generator can refine proofs based on external feedback, it fails to evaluate its own work with the same rigor as the dedicated verifier.

This observation motivated us to endow the proof generator with genuine verification capabilities.
During training, we prompt the generator $\pi_\theta$ to produce a proof $Y$ followed by a self-analysis $Z$ that follows the same format and rubrics $\mathcal{I}_v$ as the verifier (see Appendix~\ref{app:proof_generation}).
We denote the proof score predicted in the self-analysis as $s^\prime$.

To ensure faithful self-evaluation, we use the verifier $\pi_\phi$ to assess both components: the proof $Y$ receives score $R_Y = s$, and the self-analysis $Z$ receives a meta-verification score $R_\text{meta}(Z) = ms$.
The reward function combines these assessments:

\begin{align}
    R &= R_{\text{format}}(Y, Z) \cdot (\alpha \cdot R_Y + \beta \cdot R_Z) \\
    R_Z &= R_{\text{score}}(s^\prime, s) \cdot R_{\text{meta}}(Z)
\end{align}
where $R_{\text{format}}(Y, Z)$ verifies that both the proof and self-analysis follow the specified format, $R_{\text{score}}(s^\prime, s)$ rewards accurate self-assessment.
We set $\alpha = 0.76$ and $\beta = 0.24$.
This reward structure creates the following incentives:
\begin{itemize}
    \item Faithful acknowledgment of errors is rewarded over false claims of correctness.
    \item The highest rewards come from producing correct proofs and accurately recognizing their rigor.
    \item A good strategy to obtain high rewards for the proof generator is to identify and resolve as many issues as possible before finalizing the response.
\end{itemize}

\subsection{Synergy Between Proof Verification and Generation}

The proof verifier and generator create a synergistic cycle:
the verifier improves the generator, and as the generator improves, it produces new proofs that challenge the verifier's current capabilities.
These challenging cases -- where the verifier may fail to identify issues in a single attempt -- become valuable training data for enhancing the verifier itself.

To retrain and improve the verifier, we need labeled correctness data for newly generated proofs.
Manual annotation, while straightforward, becomes increasingly time-consuming as problems grow harder and errors become more subtle.
To boost annotation efficiency, we generated multiple verifier analyses per proof to surface potential issues for human review.

From this AI-assisted annotation process, we recognized two facts that make it feasible to push the level of automation a step further:

\begin{enumerate}
    \item Scaling verifier samples increases the probability of catching real issues in flawed proofs.
    \item Reviewing the verifier's identified issues is exactly \textbf{meta-verification}, which is easier than identifying issues from scratch. Meta-verification is also more sample-efficient for LLMs to master.
\end{enumerate}
Building on these observations, we developed the following automated labeling process:

\begin{enumerate}
    \item For each proof, generate $n$ independent verification analyses
    \item For analyses reporting issues (scores 0 or 0.5), generate $m$ meta-verification assessments to validate the identified problems.
    An analysis is deemed valid if the majority of meta-assessments confirm its findings
    \item For each proof, we examine analyses that assign the lowest score.
    If at least $k$ such analyses are deemed valid, the proof is labeled with that lowest score.
    If no legitimate issues are identified across all verification attempts, the proof is labeled with 1.
    Otherwise, the proof is discarded or routed to human experts for labeling
\end{enumerate}
In our last two training iterations, this fully automated pipeline replaced human annotation entirely.
Quality checks confirmed that the automated labels aligned well with expert judgments. 
\section{Experiments}

\begin{figure}[h]
  \centering
  \includegraphics[width=0.99\textwidth]{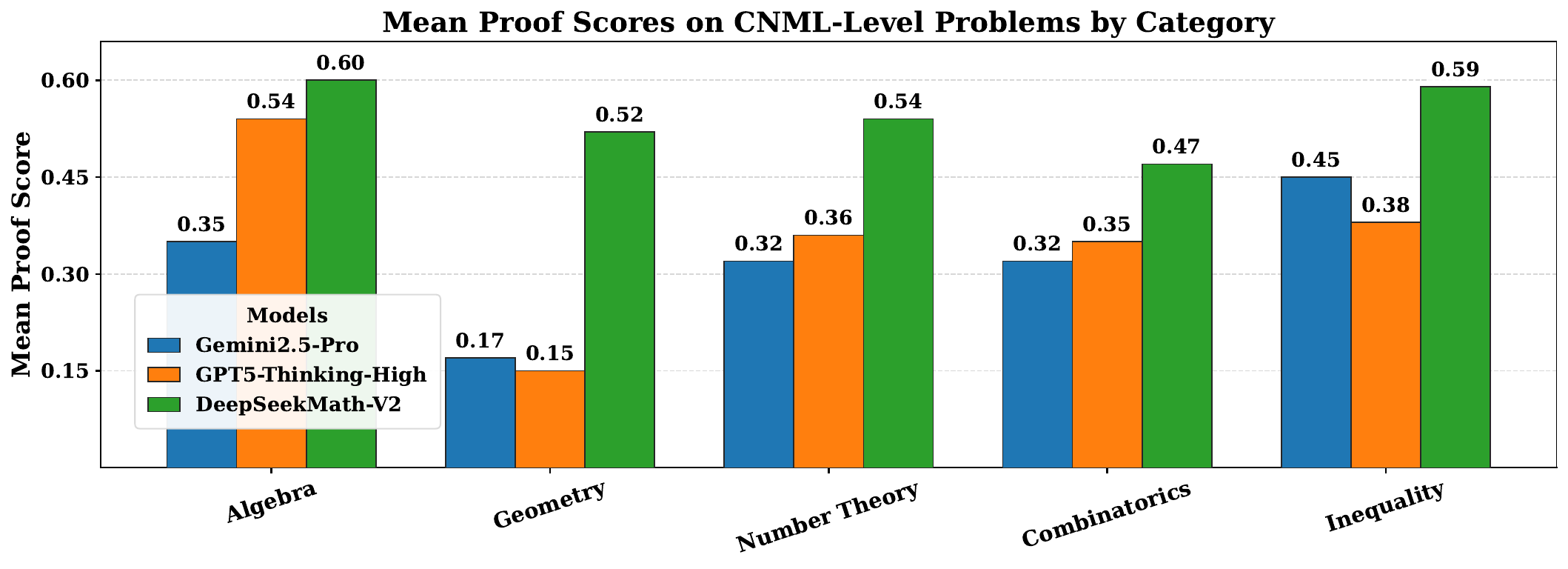}
  \caption{Average proof scores on CNML-level problems by category and model, as evaluated by our verifier.}
  \label{fig:one-shot}
\end{figure}

\subsection{Training Settings}

We employed Group Relative Policy Optimization (GRPO) \citep{deepseekmath} for reinforcement learning, iteratively optimizing proof verification and generation capabilities as described in Section~\ref{sec:method}.
In each iteration, we first optimized proof verification. The proof generator was then initialized from the verifier checkpoint and optimized for proof generation.
Starting from the second iteration, the proof verifier was initialized with a checkpoint that consolidated both verification and generation capabilities from the previous iteration through rejection fine-tuning.

\subsection{Evaluation Benchmarks}

We evaluate our final proof generator on the following theorem proving benchmarks:

\noindent
\textbf{In-House CNML-Level Problems} 91 theorem-proving problems spanning algebra (13), geometry (24), number theory (19), combinatorics (24), and inequality (11), comparable in difficulty to problems from Chinese National High School Mathematics League (CNML)

\noindent
\textbf{Competition Problems}
\begin{itemize}
    \item \textbf{IMO 2025} (6 problems): The International Mathematical Olympiad, the premier global mathematics competition for pre-university students
    \item \textbf{CMO 2024} (6 problems): The China Mathematical Olympiad, China's national championship
    \item \textbf{Putnam 2024} (12 problems): The William Lowell Putnam Competition, the preeminent mathematics competition for undergraduate students in North America
    \item \textbf{ISL 2024} (31 problems): The IMO Shortlist, a collection of problems proposed by participating countries and considered by the Problem Selection Committee for potential inclusion in IMO 2024
    \item \textbf{IMO-ProofBench} (60 problems): Developed by the DeepMind team behind DeepThink IMO-Gold \citep{deepthinkimo}, this benchmark \citep{imobench} is divided into a basic set (30 problems, pre-IMO to IMO-Medium difficulty) and an advanced set (30 challenging problems simulating complete IMO examinations, up to IMO-Hard level)
\end{itemize}

\subsection{Evaluation Results}

\subsubsection{One-Shot Generation}

We first evaluate the model's ability to generate correct proofs without iterative refinement.
On the in-house problems, we generated 8 proof samples per problem for each evaluated model.
Proof correctness was measured by majority voting across 8 verification analyses produced by our final verifier.
As shown in Figure~\ref{fig:one-shot}, across all categories of CNML-level problems -- algebra, geometry, number theory, combinatorics, and inequality -- DeepSeekMath-V2 consistently outperforms GPT-5-Thinking-High \citep{gpt5} and Gemini 2.5-Pro \citep{gemini}, demonstrating superior theorem-proving ability across domains.

\subsubsection{Sequential Refinement with Self-Verification}

\begin{wrapfigure}{r}{0.52\textwidth}  
  \centering
  \vspace{-6pt}
  \includegraphics[width=0.5\textwidth]{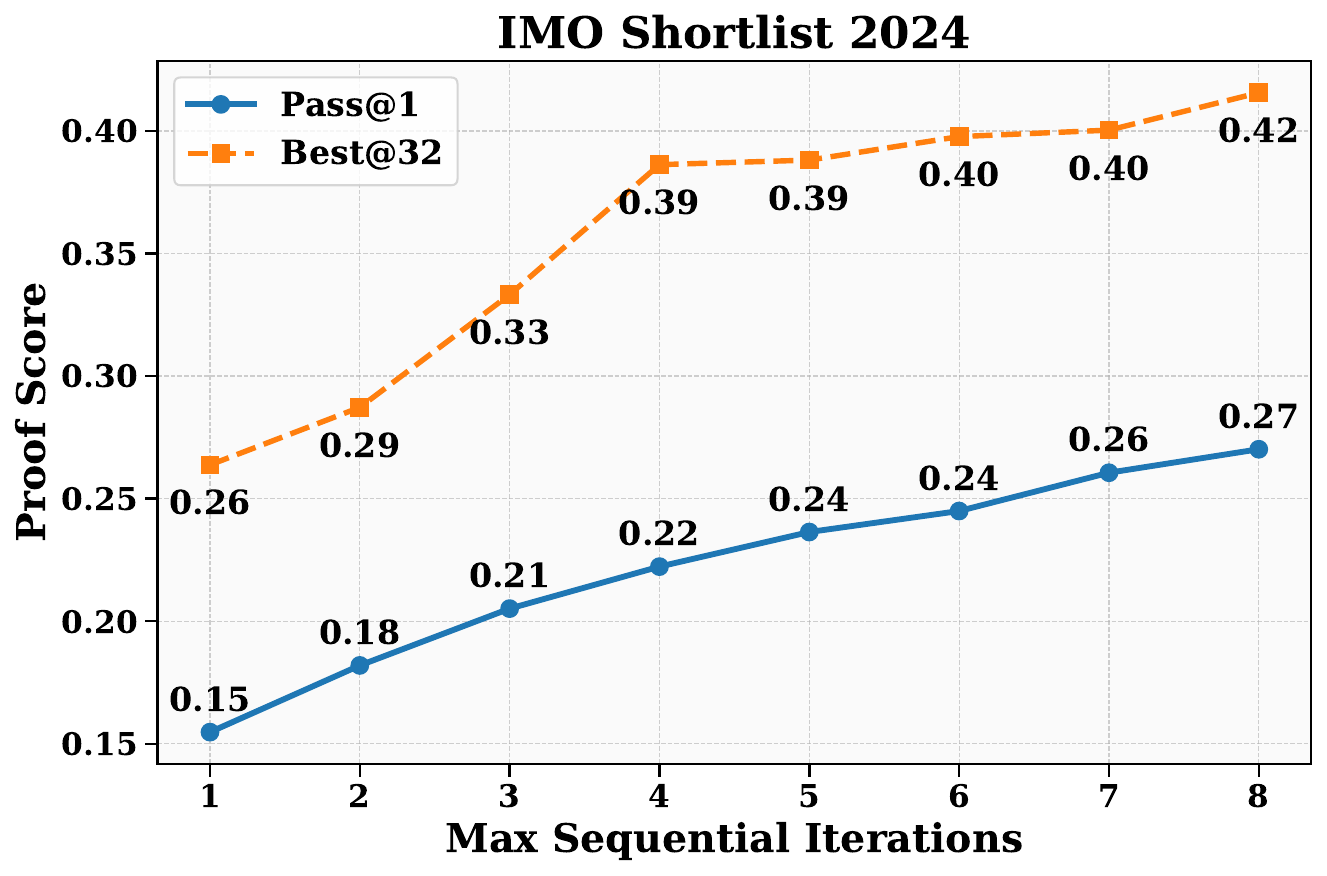} 
  \caption{Proof quality improvements as the maximum sequential iterations varies from 1 (no refinement) to 8 (initial generation plus up to 7 refinements based on self verification).\label{fig:seq_refine}}
  \vspace{-6pt}
\end{wrapfigure}

For challenging problems from competitions like IMO and CMO, models often cannot generate comprehensive and rigorous proofs in a single attempt within the 128K token limit.
When this occurs, our proof generator recognizes its proof is invalid through self-verification but lacks the context length to resolve all identified issues in a single attempt.

To explore how extended context and self-verification can improve proof quality, we evaluate sequential refinement with self-verification.
This approach first generates a proof with self-analysis, then iteratively re-prompts the generator with its previous output (see Appendix \ref{app:proof_refinement} for the refinement prompt), allowing it to address identified issues.
The process continues until the generator assigns itself a perfect score or reaches the maximum number of sequential attempts.

Figure~\ref{fig:seq_refine} demonstrates proof quality improvement through sequential refinement on IMO Shortlist 2024 problems.
For each problem, we launched 32 independent refinement threads.
Proof correctness was measured by majority voting across 32 verification analyses from our final verifier.
We report two metrics in Figure~\ref{fig:seq_refine}: (1) Pass@1 -- the average score of the final proof from each thread, and (2) Best@32 -- the score of the best proof per problem, selected by self-assigned scores across all threads.
The self-selected best proofs achieve significantly higher verification scores than the thread average, demonstrating our generator's ability to accurately assess proof quality.
Furthermore, Pass@1 improves substantially as maximum sequential attempts increase, showing that self-verification effectively guides iterative improvement.
These results confirm that our generator can reliably differentiate between high-quality and flawed proofs, and leverage this self-awareness to systematically improve its mathematical reasoning.

\begin{figure}[h]
  \centering
  \includegraphics[width=0.97\textwidth]{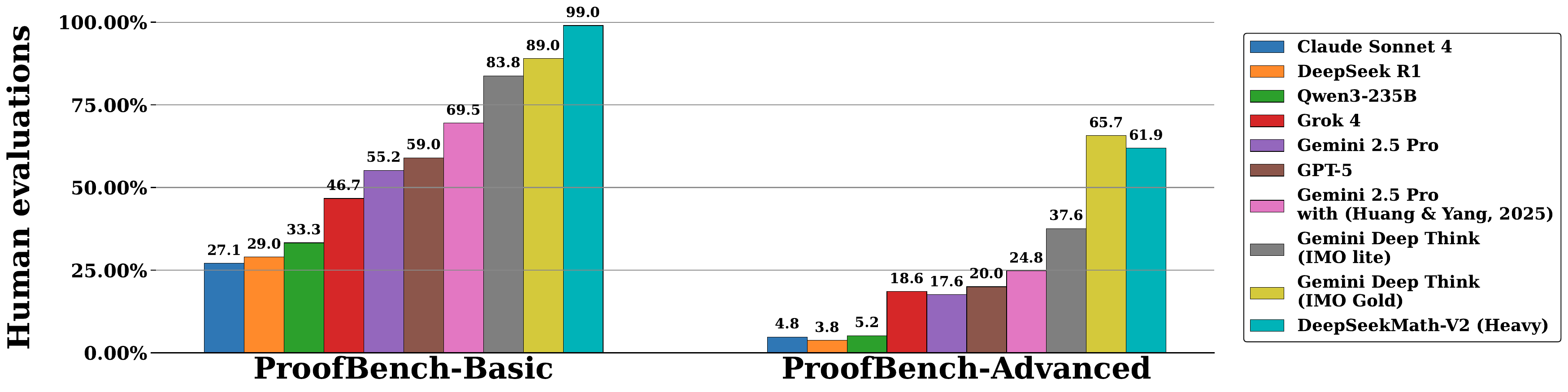}
  \vspace{-0.1in}
  \caption{Expert evaluation results on the Basic and Advanced subsets of IMO-ProofBench. All results are sourced from \cite{imobench}, with the exception of DeepSeekMath-V2, which was evaluated by our experts following the grading guidelines.}
  \label{fig:imo_proof_bench}
\end{figure}

\subsubsection{High-Compute Search}

\begin{wraptable}{r}{0.52\textwidth}
  \centering
  \begin{tabular}{lll}
    \toprule
      Contest & Problems & Points \\
    \midrule
      IMO 2025 & \textbf{\colorbox{mygray}{P1}}, \textbf{\colorbox{mygray}{P2}}, \textbf{\colorbox{mygray}{P3}}, \textbf{\colorbox{mygray}{P4}}, \textbf{\colorbox{mygray}{P5}} & 83.3\% \\
    \midrule
      CMO 2024 & \textbf{\colorbox{mygray}{P1}}, \textbf{\colorbox{mygray}{P2}}, \textbf{\colorbox{mygray}{P4}}, \textbf{\colorbox{mygray}{P5}}, \textbf{\underline{P6}} & 73.8\% \\
    \midrule
      Putnam 2024 & \textbf{\colorbox{mygray}{A1}} $\sim$ \textbf{\colorbox{mygray}{B4}}, \textbf{\underline{B5}}, \textbf{\colorbox{mygray}{B6}} & 98.3\% \\
    \bottomrule
  \end{tabular}
  \caption{Problems in gray are \textbf{\colorbox{mygray}{fully solved}}, while underlined problems received \textbf{\underline{partial credit}}. \label{tab:search}}
\end{wraptable}

To solve the most challenging problems, we scaled both verification and generation compute -- using extensive verification to identify subtle issues and parallel generation to explore diverse proof strategies.

Our approach maintains a pool of candidate proofs for each problem, initialized with 64 proof samples with 64 verification analyses generated for each.
In each refinement iteration, we select the 64 highest-scoring proofs based on average verification scores and pair each with 8 randomly selected analyses, prioritizing those identifying issues (scores 0 or 0.5).
Each proof-analysis pair is used to generate one refined proof, which then updates the candidate pool.
This process continues for up to 16 iterations or until a proof successfully passes all 64 verification attempts, indicating high confidence in correctness.
All experiments used a single model, our final proof generator, which performs both proof generation and verification.

To validate our results, mathematical experts assessed the highest-scoring proofs.
As shown in Table~\ref{tab:search}, our approach solved 5 of 6 problems from IMO 2025 and 4 problems plus partial credit on another from CMO 2024, achieving gold medal performance in both pinnacle high-school competitions.
On Putnam 2024, the preeminent undergraduate mathematics competition, our model solved 11 of 12 problems completely and the remaining problem with minor errors, scoring 118/120 and surpassing the highest human score of 90.
Figure \ref{fig:imo_proof_bench} shows the results on IMO-ProofBench.
Our approach outperforms DeepMind's DeepThink (IMO Gold) on the basic set and remains competitive on the advanced set, while substantially outperforming all other baselines.
We observe that the hardest IMO-level problems remain challenging for our model.
Notably, for problems not fully solved, our generator typically identifies the genuine issues in its proofs, while fully solved problems pass all 64 verification attempts.
This demonstrates that we can successfully train LLM-based verifiers to assess proofs previously considered difficult to verify automatically.
By scaling test-time compute under verifier guidance, our model solves problems that require hours of effort from human competitors.

\section{Related Work}

Reasoning models \citep{o1,deepseek-r1} have saturated quantitative reasoning benchmarks like AIME and HMMT within one year.
This rapid advancement is partly attributed to the well-defined evaluation criterion: if we care only about final answers, then quantitative reasoning is easy to verify.
However, this final answer metric is inapplicable to theorem proving, which often requires no numerical answers but demands rigorous step-by-step derivation.
Informal mathematical proofs have long been considered hard to verify automatically, lacking reliable approaches to assess proof correctness.
Recent developments suggest this barrier may be surmountable.
Models like Gemini-2.5 Pro already demonstrate a certain level of self-verification capabilities, which can refine their own solutions to improve quality \citep{geminiimo}.
More significantly, DeepMind's internal DeepThink variant \citep{deepthinkimo} achieved gold medal performance at IMO 2025 using pure natural language reasoning -- serving as an existence proof that LLM-based verification of complex proofs is achievable.
Recent research has begun exploring whether reasoning models can evaluate proofs, both with and without reference solutions \citep{opc,imobench}, showing promising results.
In this work, we open source DeepSeekMath-V2 and our training methodology as concrete steps toward self-verifiable mathematical reasoning, showing how models can learn to verify and improve their own proofs.

Proof assistants like Lean \citep{lean} and Isabelle \citep{isabelle} offer a reliable approach to verify proofs -- proofs must be written in formal language, but once compiled, correctness is guaranteed.
AlphaProof \citep{alphaproof,alphageometry,alphageometry2}, a system specialized for formal proof search, achieved silver-level performance at IMO 2024 but required intensive computation.
While using informal reasoning to guide formal proof generation has been explored extensively \citep{dsp}, recent reasoning models have dramatically improved informal reasoning quality, making this guidance far more effective.
Systems like DeepSeek-Prover-V2 \citep{deepseekproverv2} and Seed-Prover \citep{seedprover} can now produce substantially more valid formal proofs within the same computational budget, with Seed-Prover solving 5 of 6 problems at IMO 2025.
Notably, these results were achieved without specifically optimizing the informal reasoning components for theorem proving tasks.
We believe advancing natural language theorem proving will significantly benefit formal reasoning.
We hope to contribute toward truly reliable mathematical reasoning systems that leverage both informal insights and formal guarantees to advance mathematical research.

\section{Conclusion}

We presented DeepSeekMath-V2, a model capable of both generating and verifying mathematical proofs.
By training models to identify issues in their own reasoning and incentivizing them to address these issues before finalizing outputs, we move beyond the limitations of final-answer-based rewards toward self-verifiable mathematical reasoning.
Our iterative training process -- alternating between improving verification capabilities and using these to enhance generation -- creates a sustainable cycle where each component drives the other forward.
Our key technical contributions include: (1) training an accurate and faithful LLM-based verifier for mathematical proofs, (2) using meta-verification to largely reduce hallucinated issues and ensure verification quality, (3) incentivizing the proof generator to maximize proof quality through self-verification, and (4) scaling verification compute to automatically label increasingly hard-to-verify proofs to improve the verifier without human annotation.
DeepSeekMath-V2 demonstrates strong performance on competition mathematics.
With scaled test-time compute, it achieved gold-medal scores in high-school competitions including IMO 2025 and CMO 2024, and a near-perfect score on the undergraduate Putnam 2024 competition.
This work establishes that LLMs can develop meaningful self-evaluation abilities for complex reasoning tasks.
While significant challenges remain, we hope this research direction contributes to the goal of creating self-verifiable AI systems that can solve research-level mathematics.

\bibliographystyle{abbrvnat}
\bibliography{main}

\newpage
\appendix

\section{Prompt Templates}\label{apx:cot-prompt}

\subsection{Proof Generation Prompt}
\label{app:proof_generation}

\begin{lstlisting}[frame=single]
Your task is to solve a given problem. The problem may ask you to prove a statement, or ask for an answer. If finding an answer is required, you should come up with the answer, and your final solution should also be a rigorous proof of that answer being valid.

Your final solution to the problem should be exceptionally comprehensive and easy-to-follow, which will be rated according to the following evaluation instruction:

```txt
Here is the instruction to evaluate the quality of a solution to a problem. The problem may ask for a proof of statement, or ask for an answer. If finding an answer is required, the solution should present the answer, and it should also be a rigorous proof of that answer being valid.

Please evaluate the solution and score it according to the following criteria:
- If the solution is completely correct, with all steps executed properly and clearly demonstrated, then the score is 1
- If the solution is generally correct, but with some details omitted or minor errors, then the score is 0.5
- If the solution does not actually address the required problem, contains fatal errors, or has severe omissions, then the score is 0

Additionally, referencing anything from any paper does not save the need to prove the reference. It's okay IF AND ONLY IF the solution also presents a valid proof of the reference argument(s); otherwise, if the solution omits the proof or if the proof provided is not completely correct, the solution should be scored according to the criteria above, and definitely not with a score of 1
```

In fact, you already have the ability to rate your solution yourself, so you are expected to reason carefully about how to solve a given problem, evaluate your method according to the instruction, and refine your solution by fixing issues identified until you can make no further progress.

In your final response, you should present a detailed solution to the problem followed by your evaluation of that solution.
- To give a good final response, you should try your best to locate potential issues in your own (partial) solution according to the evaluation instruction above, and fix them as many as you can.
- A good final response should just faithfully present your progress, including the best solution you can give, as well as a faithful evaluation of that solution.
- Only when you fail to locate any issues in your solution should you score it with 1.
- If you do notice some issues in your solution but fail to resolve them with your best efforts, it's totally ok to faithfully present the issues in your final response.
- The worst final response would provide a wrong solution but lie that it's correct or claim that it's correct without careful error checking. A better version should faithfully identify errors in the solution. Remember! You CAN'T cheat! If you cheat, we will know, and you will be penalized!

Your final response should be in the following format:

## Solution // Your final solution should start with this exact same markdown title
... // Your final solution to the problem here. You should try your best to optimize the quality of your solution according to the evaluation instruction above before finalizing it here.

## Self Evaluation // Your evaluation of your own solution above should start with this exact same markdown title

Here is my evaluation of the solution: // Your analysis should start with this exact same phrase
... // Your evaluation here. You are required to present in detail the key steps of the solution or the steps for which you had doubts regarding their correctness, and explicitly analyze whether each step is accurate: for correct steps, explain why you initially doubted their correctness and why they are indeed correct; for erroneous steps, explain the reason for the error and the impact of that error on the solution. You should analyze your solution faithfully. E.g., if there are issues in your final solution, you should point it out.

Based on my evaluation, the final overall score should be:
\\boxed{{...}} // where ... should be the final overall score (0, 0.5, or 1, and nothing else) based on the evaluation instruction above. You should reach this score ONLY AFTER careful RE-examination of your own solution above

---

Here is your task input:

## Problem
{question}
\end{lstlisting}

\subsection{Proof Verification Prompt}
\label{app:proof_verification}
\begin{lstlisting}[frame=single]
## Instruction

Your task is to evaluate the quality of a solution to a problem. The problem may ask for a proof of statement, or ask for an answer. If finding an answer is required, the solution should present the answer, and it should also be a rigorous proof of that answer being valid.

Please evaluate the solution and score it according to the following criteria:
- If the solution is completely correct, with all steps executed properly and clearly demonstrated, then the score is 1
- If the solution is generally correct, but with some details omitted or minor errors, then the score is 0.5
- If the solution does not actually address the required problem, contains fatal errors, or has severe omissions, then the score is 0
- Additionally, referencing anything from any paper does not save the need to prove the reference. It's okay IF AND ONLY IF the solution also presents a valid proof of the reference argument(s); otherwise, if the solution omits the proof or if the proof provided is not completely correct, the solution should be scored according to the criteria above, and definitely not with a score of 1

Please carefully reason out and analyze the quality of the solution below, and in your final response present a detailed evaluation of the solution's quality followed by your score. Therefore, your response should be in the following format:

Here is my evaluation of the solution:
... // Your evaluation here. You are required to present in detail the key steps of the solution or the steps for which you had doubts regarding their correctness, and explicitly analyze whether each step is accurate: for correct steps, explain why you initially doubted their correctness and why they are indeed correct; for erroneous steps, explain the reason for the error and the impact of that error on the solution.

Based on my evaluation, the final overall score should be:
\\boxed{{...}} // where ... should be the final overall score (0, 0.5, or 1, and nothing else) based on the above criteria

---

Here is your task input:

## Problem
{question}

## Solution
{proof}
\end{lstlisting}

\subsection{Meta-Verification Prompt}
\label{app:meta_verification}

\begin{lstlisting}[frame=single]
You are given a "problem", "solution", and "solution evaluation", and you need to assess the whether this "solution evaluation" is reasonable.

First, "solution evaluation" is generated to evaluate the quality of the "solution", by prompting a verifier with the rules below (these are not your rules):

```
Please evaluate the solution and score it according to the following criteria:
- If the solution is completely correct, with all steps executed properly and clearly demonstrated, then the score is 1
- If the solution is generally correct, but with some details omitted or minor errors, then the score is 0.5
- If the solution does not actually address the required problem, contains fatal errors, or has severe omissions, then the score is 0

Additionally, referencing anything from any paper does not save the need to prove the reference. It's okay IF AND ONLY IF the solution also presents a valid proof of the reference argument(s); otherwise, if the solution omits the proof or if the proof provided is not completely correct, the solution should be scored according to the criteria above, and definitely not with a score of 1
```

Next, I will introduce the rules for you to analyze the quality of the "solution evaluation":

1. Your task is to analyze the "solution evaluation". You do not need to solve the "problem", nor do you need to strictly assess whether the "solution" is accurate. Your only task is to strictly follow the rules below to evaluate whether the "solution evaluation" is reasonable.

2. You need to analyze the content of the "solution evaluation" from three aspects:

Step Restatement: In the "solution evaluation", certain behaviors of the "solution" may be restated. You need to return to the original text of the "solution" and check whether the "solution" actually has these behaviors mentioned in the "solution evaluation".

Defect Analysis: "solution evaluation" may point out errors or defects in the "solution". You need to carefully analyze whether the mentioned errors and defects are indeed valid.

Expression Analysis: Whether the "solution evaluation"'s expressions are accurate.

Score Analysis: Whether the final score given by the "solution evaluation" matches the defects it found. You need to analyze according to the scoring rules given above.

3. The most important part is **defect analysis**: In this part, your core task is to check whether the errors or defects of the "solution" pointed out in the "solution evaluation" are reasonable. In other words, any positive components about the "solution" in the "solution evaluation", regardless of whether they are reasonable, are not within your evaluation scope.

- For example: If the "solution evaluation" says that a certain conclusion in the "solution" is correct, but actually this conclusion is incorrect, then you do not need to care about this point. All parts that the "solution evaluation" considers correct do not belong to your evaluation scope.
- Specifically: If the "solution evaluation" believes that the "solution" is completely accurate and has not found any errors or defects, then regardless of whether the "solution" itself is actually accurate, even if there are obvious errors, you should still consider its analysis of errors to be reasonable.

**Importantly**, for defects found by the "solution evaluation", you need to analyze two points simultaneously:

- whether this defect actually exists
- whether the "solution evaluation"'s analysis of this defect is accurate

These two aspects constitute the analysis of defects.

4. About **expression analysis**, if there are certain expression errors in the "solution evaluation", even minor errors in details, you need to identify them. However, please note that identifying incorrect steps in the "solution" as correct steps does not constitute an **expression error**.

In practice, expression errors include but are not limited to:

- If the "solution evaluation" identifies some reasoning step(s) in the "solution" as incorrect, then it cannot further indicate that subsequent conclusion(s) depending on those reasoning step(s) are wrong, but can only indicate that subsequent conclusion(s) are "not rigorously demonstrated."
- Typos and calculation errors made by "solution evaluation"
- Inaccurate restatement of content from "solution"

5. Finally, you need to present your analysis of the "solution evaluation" in your output and also rate its quality based on the rules below:

First, if there is at least one unreasonable defect among the defects found by the "solution evaluation", then you only need to do **defect analysis**:

- If all defects found by the "solution evaluation" are unreasonable, then you should rate it with \(0\)
- If some defects found by the "solution evaluation" are reasonable and some are unreasonable, then your rating should be \(0.5\)

Next, if the "solution evaluation" points out no errors or defects, or all defects found by the evaluation are reasonable, then you should do the following things:

- Analyze whether "expression errors" exist in the "solution evaluation" (**expression analysis**) or whether "solution evaluation" gives a wrong score according to the rules for "solution evaluation" (**score analysis**). If yes, you should rate the "solution evaluation" with \(0.5\); if no, your rating should be \(1\)

Your output should follow the format below:

Here is my analysis of the "solution evaluation":
... // Your analysis here.

Based on my analysis, I will rate the "solution evaluation" as:
\\boxed{{...}} // where ... should be a numerical rating of the "solution evaluation" (0, 0.5, or 1, and nothing else) based on the criteria above.

---

Here is your task input:

## Problem
{question}

## Solution
{proof}

## Solution Evaluation
{proof analysis}
\end{lstlisting}

\subsection{Proof Refinement Prompt}
\label{app:proof_refinement}

\begin{lstlisting}[frame=single]
{proof_generation_prompt}

## Candidate Solution(s) to Refine
Here are some solution sample(s) along with their correctness evaluation(s). You should provide a better solution by solving issues mentioned in the evaluation(s), or by re-using promising ideas mentioned in the solution sample(s), or by doing both.

{proof}
{proof analyses}

## Final Instruction
Your final response should follow the format above, including a `## Solution` section followed by a `## Self Evaluation` section
\end{lstlisting}

\end{CJK*}
\end{document}